\documentclass[11pt, a4paper, logo, onecolumn, copyright, numbering]{googledeepmind}

\pdftrailerid{redacted}

\makeatletter
\renewcommand\bibentry[1]{\nocite{#1}{\frenchspacing\@nameuse{BR@r@#1\@extra@b@citeb}}}
\makeatother

\usepackage{kantlipsum, lipsum}
\usepackage{dsfont}
\usepackage{gdm-colors}
\usepackage{adjustbox}
\usepackage{amsmath}
\usepackage{amssymb}
\usepackage{bm}
\usepackage{colortbl}
\usepackage{comment}
\usepackage{dsfont}
\usepackage{enumitem}
\usepackage{graphicx} 
\usepackage{mathtools}
\usepackage{multicol}
\usepackage{multirow}
\usepackage{lscape}
\usepackage{scalerel}
\usepackage{soul}
\usepackage{subcaption}
\usepackage{tabularx}
\usepackage[most]{tcolorbox}
\usepackage{xspace}
\usepackage{wrapfig}

\usepackage[authoryear, sort&compress, round]{natbib}

\graphicspath{{figures/}}

\title{Foundational Autoraters: 
\\ Taming Large Language Models for \\ Better Automatic Evaluation}
\correspondingauthor{ttvu@google.com and kalpeshk@google.com}

\reportnumber{001} %

\renewcommand{\today}{2024-07-15}

\author[*,1]{Tu Vu}
\author[*,2]{Kalpesh Krishna}
\author[3]{\\ Salaheddin Alzubi}
\author[1]{Chris Tar}
\author[2]{Manaal Faruqui}
\author[1]{Yun-Hsuan Sung}

\affil[*]{Co-lead (equal contribution)}
\affil[1]{Google DeepMind}
\affil[2]{Google}
\affil[3]{UMass Amherst}

\newcommand{\namedref}[2]{\hyperref[#2]{#1~\ref*{#2}}}

\newcommand{\sectionref}[1]{Section \ref{#1}}
\newcommand{\tableref}[1]{\namedref{Table}{#1}}
\newcommand{\figureref}[1]{\namedref{Figure}{#1}}
\newcommand{\appendixref}[1]{\namedref{Appendix}{#1}}

\newcommand{\ssc}[1]{{\small \sc #1}\xspace}

\newcommand{\llm}{{\ssc{LLM}}\xspace}

\newcommand{\smallurl}[1]{\begin{small}\url{#1}\end{small}}

\begin{abstract}
As large language models (LLMs) advance, it becomes more challenging to reliably evaluate their output due to the high costs of human evaluation. To make progress towards better LLM autoraters, we introduce \textcolor{blue}{FLAMe}, a family of \textcolor{blue}{F}oundational \textcolor{blue}{L}arge \textcolor{blue}{A}utorater \textcolor{blue}{M}od\textcolor{blue}{e}ls. FLAMe is trained on our large and diverse collection of 100$+$ quality assessment tasks comprising 5M$+$ human judgments, curated and standardized using \emph{publicly released human evaluations} from previous research. FLAMe significantly improves generalization to a wide variety of held-out tasks, outperforming LLMs trained on proprietary data like GPT-4 and Claude-3 on many tasks. We show that FLAMe can also serve as a powerful starting point for further downstream fine-tuning, using reward modeling evaluation as a case study (FLAMe-RM). Notably, on RewardBench, our FLAMe-RM-24B model (with an accuracy of 87.8\%) is the top-performing generative model trained exclusively on permissively licensed data, outperforming both GPT-4-0125 (85.9\%) and GPT-4o (84.7\%). Additionally, we explore a more computationally efficient approach using a novel \textit{tail-patch fine-tuning} strategy to optimize our FLAMe multitask mixture for reward modeling evaluation (FLAMe-Opt-RM), offering competitive RewardBench performance while requiring approximately 25$\times$ less training datapoints. Overall, our FLAMe variants outperform all popular proprietary LLM-as-a-Judge models we consider across 8 out of 12 autorater evaluation benchmarks, encompassing 53 quality assessment tasks, including RewardBench and LLM-AggreFact. Finally, our analysis reveals that FLAMe is significantly less biased than these LLM-as-a-Judge models on the CoBBLEr autorater bias benchmark, while effectively identifying high-quality responses for code generation. %
\vspace{-0.08in}
\end{abstract}

\begin{document}

\maketitle

\section{Introduction}

\begin{figure*}[h]
\centering
\includegraphics[width=0.99\textwidth]{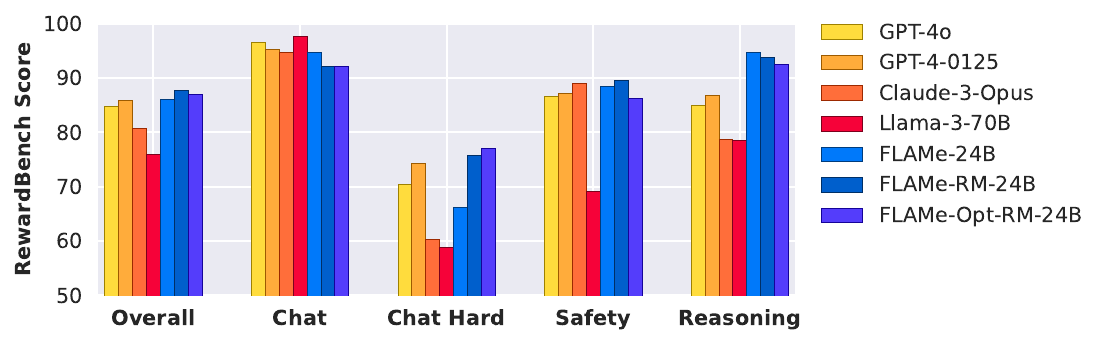}
\caption{Our FLAMe-24B variants outperform popular proprietary LLM-as-a-Judge models like GPT-4 and Claude-3 on many held-out autorater evaluation benchmarks, including RewardBench. Notably, FLAMe-RM, with an overall accuracy of 87.8\%, is the top-performing generative model trained solely on permissively licensed data on RewardBench, surpassing both GPT-4-0125 (85.9\%) and GPT-4o (84.7\%).}
\label{fig:rewardbench_performance}
\vspace{-0.15in}
\end{figure*}

The increasing power and versatility of large language models (LLMs) bring with them a growing challenge: \textit{How can we reliably evaluate their long-form outputs?} Recent research suggests a promising solution: these models themselves, after undergoing large-scale multitask instruction tuning, can generalize to follow new human instructions~\citep{mishra-etal-2022-cross,wei-etal-2022-finetuned,sanh-etal-2022-multitask,longpre-etal-2023-the,chung-etal-2024-scaling}, making them suitable for use as autoraters of model outputs. This is particularly appealing because human evaluation, though crucial for assessing model performance, is limited by subjectivity~\citep{krishna-etal-2023-longeval}, variability among raters~\citep{karpinska-etal-2021-perils}, and the high costs of extensive evaluations~\citep{min-etal-2023-factscore,vu-etal-2023-fresh,wei-etal-2024-long}.

To align LLM autoraters with human preferences, training on human judgments is crucial~\citep{ouyang-etal-2022-training}. However, obtaining these judgments is both costly and time-consuming. Collecting existing human evaluations from previous research seems promising but faces challenges such as lack of standardization, diverse evaluation criteria, inadequate documentation, data privacy, and proprietary concerns. Alternatively, using model outputs for autorater training offers consistency~\citep{jiang-etal-2024-tigerscore,kim-etal-2024-prometheus-2} but also carries with risks, including reinforcing biases and hallucinations~\citep{gudibande-etal-2023-the,muennighoff-etal-2023-octopack}. Additionally, it may violate terms of use for proprietary LLM services, which prohibit using their models' outputs to develop competing models.\footnote{\smallurl{https://openai.com/policies/terms-of-use, https://policies.google.com/terms/generative-ai}
}

To address these limitations, we curated and standardized human evaluations from prior research to create FLAMe, a collection of 102 quality assessment tasks comprising more than 5.3M total human judgments (\sectionref{sec:flame-modeling}). FLAMe spans a wide variety of task types, from assessing machine translation quality to evaluating how well AI assistants follow user instructions. We hypothesized that training on this large and diverse data collection would enable LLM autoraters to learn robust, generalized patterns of human judgment, minimizing the impact of noisy or low-quality human judgments.

For transparency and reproducibility, we use only \emph{publicly available human evaluation data with permissive licenses} from previous studies (\sectionref{sec:principles-training-data}). To overcome challenges in collecting such data, which rarely adhere to a particular standard and often lack documentation, we thoroughly examined the associated research (\sectionref{sec:unified-format}) and additionally consulted with the original authors to address ambiguities or inconsistencies (spending 3-4 hours per dataset).

\begin{figure*}[t]
\centering
\includegraphics[width=0.94\textwidth]{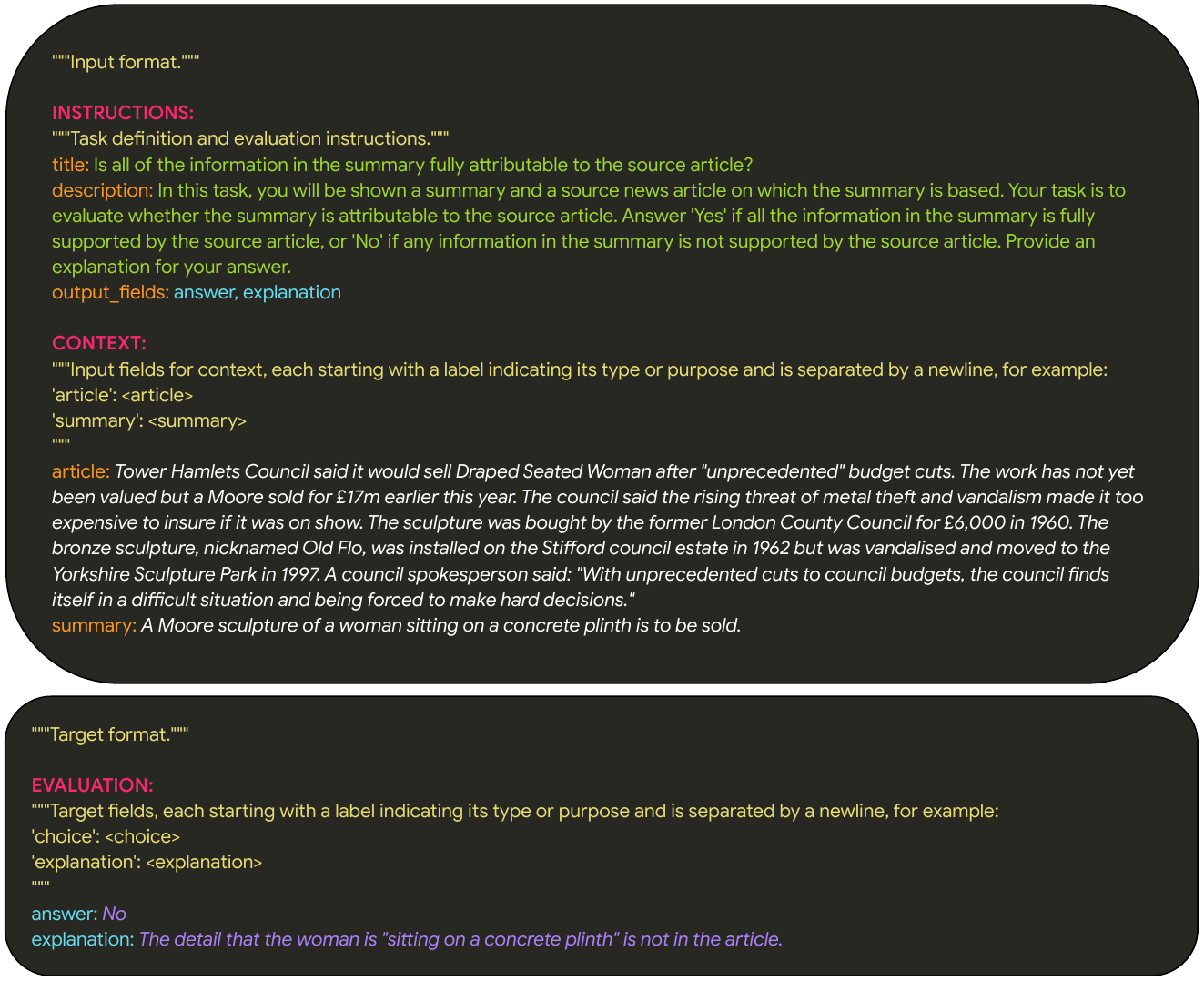}
\caption{All our quality assessment tasks are formulated into a unified \textit{text-to-text} format with manually crafted task definitions and evaluation instructions. We format training examples as input-target pairs, where the input includes task-specific context, and the target contains the expected human evaluations.}

\label{fig:unified_task_format}
\vspace{-0.15in}
\end{figure*}

We train LLM autoraters using supervised, multitask fine-tuning on our data collection. Inspired by T5's unified task format~\citep{raffel-etal-2020-exploring}, we convert all our quality assessment tasks into a \textit{text-to-text} format with manually crafted task definitions and evaluation instructions. All training examples are formulated as input-target pairs, where the input includes task-specific context, and the target contains the expected human evaluations (see~\figureref{fig:unified_task_format}). This approach facilitates effective transfer learning across tasks, enabling our models to interpret and respond to various tasks consistently. Additionally, our task format is simple, intuitive, and easily accommodates new tasks.

Our approach can be viewed as developing general-purpose LLM autoraters that can perform various quality assessment tasks. We demonstrate that training an instruction-tuned LLM, i.e., PaLM-2-24B~\citep{anil-etal-2023-palm}, on our FLAMe collection significantly improves generalization to a wide variety of held-out tasks, outperforming models like GPT-4, Claude-3, and Llama-3 on many tasks. This suggests that our large-scale multitask instruction tuning approach effectively equips the model with general-purpose quality assessment capabilities.

Motivated by these results, we further investigate the impact of using FLAMe as a powerful starting point for fine-tuning on targeted downstream applications, using reward modeling evaluation as a case study (FLAMe-RM). Specifically, we further fine-tune FLAMe for only 50 steps on a mixture of four datasets with human pairwise preference judgments, covering chat, reasoning, and safety. Our resulting FLAMe-RM-24B model significantly improves FLAMe's performance on RewardBench~\citep{lambert-etal-2024-rewardbench}, achieving an overall accuracy of 87.8\% (up from 86.0\%). Notably, it is \textit{the top-performing generative model trained solely on permissively licensed data}, outperforming both GPT-4-0125 (85.9\%) and GPT-4o (84.7\%); see~\figureref{fig:rewardbench_performance}.

Additionally, we present FLAMe-Opt-RM, a computationally efficient method that optimizes our FLAMe multitask mixture for targeted reward modeling evaluation. Using a novel \textit{tail-patch fine-tuning} technique, we analyze the impact of each dataset on specific RewardBench distributions, allowing us to determine the optimal proportions of individual datasets in our multitask mixture.  By fine-tuning the initial instruction-tuned PaLM-2-24B checkpoint on this optimized mixture for only 5000 steps, we obtain competitive RewardBench performance (87.0\%) compared to FLAMe (86.0\%), using approximately 25$\times$ less training datapoints.

Overall, our FLAMe variants outperform all popular proprietary LLM-as-a-Judge models we consider across 8 out of 12 autorater evaluation benchmarks (1 held-in and 11 held-out), encompassing 53 quality assessment tasks, including benchmarks like RewardBench and LLM-AggreFact~\citep{tang-etal-2024-minicheck}.

Finally, we investigate whether biases exist in our LLM autoraters, a common criticism of LLM-as-a-Judge autoraters (\sectionref{sec:autorater-bias}), and their potential utility for AI development, particularly in identifying high-quality model responses (\sectionref{sec:reranker-results}). Our analysis results reveal that FLAMe variants are significantly less biased than popular LLM-as-a-Judge models on the CoBBLEr autorater bias benchmark~\citep{koo-etal-2023-benchmarking}, showing more robustness to changes in pairwise ordering, response length, and irrelevant context. Additionally, we find that FLAMe effectively re-ranks LLM responses to Python programming prompts in the HumanEval benchmark~\citep{chen-etal-2021-evaluating}, improving pass@1 by 6-10\% across settings.

In summary, our main contributions are: 
\begin{enumerate}
    \item \textbf{Data Collection}: We curated and standardized human evaluations from permissively licensed datasets to create a collection of 100+ diverse quality assessment tasks comprising 5M$+$ human judgments. To facilitate future research, we will make our data collection publicly available.
    \item \textbf{LLM Autoraters}: We demonstrate the effectiveness of using our multitask mixture both in training general-purpose LLM autoraters (FLAMe) and optimizing LLM autoraters for targeted downstream applications (FLAMe-RM and FLAMe-Opt-RM). Our LLM autoraters outperform all popular proprietary LLM-as-a-Judge models we consider across 8 out of 12 autorater evaluation benchmarks, covering 53 quality assessment tasks, including RewardBench and LLM-AggreFact.
    \item \textbf{Computationally Efficient Multitask Training}: We introduce a computationally efficient method using a novel fine-tuning strategy to optimize our multitask mixture for targeted distributions, achieving competitive performance with significantly less compute.
\end{enumerate}

Our work demonstrates the potential of accessible AI solutions, which we hope will spur more fundamental research into reusable human evaluations and the development of effective and efficient LLM autoraters.

\section{Related Work}
\label{sec:related_work}
Below, we discuss existing literature in the space of autoraters, drawing connections to FLAMe.

\paragraph{Automatic Evaluation Metrics:} %
Traditional metrics like BLEU~\citep{papineni-etal-2002-bleu} and ROUGE~\citep{lin-etal-2024-rouge} assess the lexical overlap between model output and human references. In the BERT~\citep{devlin-etal-2019-bert} era, several methods use pretrained models to measure the distributional similarity%
~\citep{zhao-etal-2019-moverscore,zhang-etal-2020-bertscore} or token probabilities~\citep{thompson-post-2020-automatic,yuan-etal-2021-bartscore}. A line of work explores statistical methods to measure the divergence between two text distributions~\citep{gehrmann-etal-2019-gltr,pillutla-etal-2021-mauve}. Other work fine-tunes pretrained models on human ratings to create automatic evaluation metrics for specific tasks, including machine translation~\citep{sellam-etal-2020-bleurt,rei-etal-2020-comet,fernandes-etal-2023-devil}%
, text summarization~\citep{durmus-etal-2020-feqa,deutsch-etal-2021-towards,goyal-durrett-2021-annotating}, question answering~\citep{chen-etal-2020-mocha,lin-etal-2022-truthfulqa}, and text simplification~\citep{maddela-etal-2023-lens}. Unlike these task-specific evaluation metrics, FLAMe is trained on various fine-grained quality assessment tasks and can be prompted during inference to tackle novel tasks.

\paragraph{LLM-as-a-Judge Autoraters:} With the advent of LLMs like ChatGPT, recent work has used these models as judges~\citep{liu-etal-2023-g,fu-etal-2024-gptscore,bai-etal-2023-benchmarking,wang-etal-2023-chatgpt,chiang-etal-2023-vicuna,chiang-lee-2023-can,bubeck-etal-2023-sparks} to evaluate \llm capabilities on various benchmarks, including AlpacaEval~\citep{li-etal-2023-alpacaeval,dubois-etal-2024-length}, MT-Bench~\citep{zheng-etal-2023-judging}, and WildBench~\citep{lin-etal-2024-wildbench}. However, LLM-as-a-Judge autoraters are often found to favor their own generated responses~\citep{liu-etal-2023-g,panickssery-etal-2024-llm,liu-etal-2023-llms,bai-etal-2023-benchmarking}, exhibiting ``cognitive'' biases towards aspects like length, order, and entity preference~\citep{koo-etal-2023-benchmarking}. In contrast, our models are trained on a large, diverse collection of human evaluations, allowing them to learn unbiased, generalized patterns of human judgment (\sectionref{sec:autorater-bias}). Unlike LLM-as-a-Judge autoraters, our models are not tasked with evaluating their own responses, avoiding self-preference bias.
\vspace{-0.5em}
\paragraph{General-purpose LLM Autoraters:} Recent work has explored training general-purpose LLM autoraters. ~\citet{jiang-etal-2024-tigerscore} introduce TIGERScore, a Llama-2 model trained on GPT-4 generated error analysis data across several tasks, including summarization, translation, data2text, long-form QA, and instruction-following. Similar approaches include InstructScore~\citep{xu-etal-2023-instructscore}, Prometheus~\citep{kim-etal-2024-prometheus}, and Prometheus-2~\citep{kim-etal-2024-prometheus-2}. Unlike these efforts, our approach relies solely on open-source human evaluations instead of model outputs. We show that FLAMe significantly outperforms Prometheus-2 on RewardBench (see \tableref{tab:rewardbench_table}).

\vspace{-0.5em}
\paragraph{Reward Models:} Our work relates to the development of reward models (RMs) used for aligning LLMs to human preferences via reinforcement learning with human feedback (RLHF)~\citep{ouyang-etal-2022-training,korbak-etal-2023-pretraining}. In RLHF, human preference data is either used to train stand-alone discriminative RMs, or directly fed into LLMs via algorithms like DPO~\citep{rafailov-etal-2024-direct} or SLiC-HF~\citep{zhao-etal-2023-slic}. While we evaluate our models as RMs in our RewardBench experiments (\sectionref{sec:experiments}), there are key distinctions: (1) RMs primarily train on pairwise preference data,\footnote{A notable exception is RL\emph{AI}F~\citep{bai-etal-2022-constitutional}, which asks the model to critique its responses based on a constitution.} whereas our models use diverse task types in a unified format; (2) RMs optimize for overall preference, while our models can be prompted to judge specific aspects of model responses (e.g., safety).

\section{The FLAMe Collection}
\label{sec:flame-modeling}
At a high level, we fine-tune instruction-tuned LLMs on our multitask mixture of standardized human evaluations (102 tasks, 5.3M human judgments). This data collection is meticulously curated to encompass human evaluations across a broad spectrum of LLM capabilities (\sectionref{sec:principles-training-data}-\ref{sec:capabilities}). We manually crafted task definitions and evaluation instructions, reformatting all tasks into a unified text-to-text format (\sectionref{sec:unified-format}).

\subsection{Task Definition}
\label{sec:task-definition}
We use the term ``task'' to refer to a specific assignment for the model, which involves presenting a text (e.g., a machine-generated summary) alongside its context (the original article) and instructing the model to evaluate one or more aspects of the text based on provided evaluation criteria (see~\figureref{fig:unified_task_format}). Each task has distinct definitions and evaluation guidelines. It is possible to derive different tasks from the same dataset. For example, HelpSteer~\citep{wang-etal-2023-helpsteer} includes human annotations for different attributes of model responses such as Helpfulness, Correctness, Coherence, Complexity, and Verbosity, allowing us to create distinct tasks, each focused on a specific attribute. Additionally, tasks with similar definitions and evaluation criteria but sourced from different datasets are treated as distinct tasks. Based on this definition, the FLAMe collection has a total of 102 distinct tasks.

\subsection{Principles for Data Collection}
\label{sec:principles-training-data}

We adhere to the following principles while choosing our datasets:

\paragraph{Public, Open-source Datasets:} For transparency and reproducibility, we use only permissively licensed datasets from HuggingFace Datasets~\citep{lhoest-etal-2021-datasets}, TensorFlow Datasets,\footnote{\smallurl{https://www.tensorflow.org/datasets}} or the original authors' GitHub repositories.

\paragraph{Human-labeled Annotations:} We exclusively use datasets with human-labeled annotations, avoiding those generated by models like GPT-4 due to potential inaccuracies and legal concerns raised in recent research~\citep{gudibande-etal-2023-the,muennighoff-etal-2023-octopack}. 
\begin{figure}[t]
\centering
\includegraphics[width=0.5\textwidth]{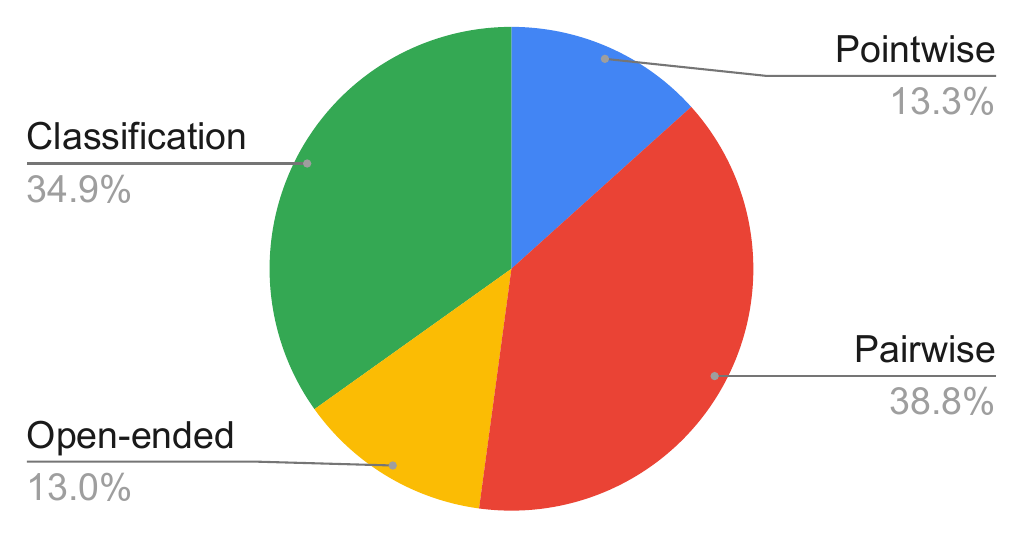}
\caption{A breakdown of our FLAMe data collection by task type, with each slice representing the percentage of datapoints (out of 5.3M) for that specific task type. More than half of FLAMe focuses on standard pairwise (\emph{``Which of the two responses is better?''}) and pointwise (\emph{``Rate the response on a Likert scale.''}) evaluation tasks. The rest of FLAMe focuses on custom classification (e.g., \emph{``Is the summary fully attributable to the source article? (Yes/No)''}) and open-ended generation (e.g., \emph{``Explain why response A is better than response B.''}) evaluation tasks.}
\label{fig:task_type_breakdown}
\end{figure}

\paragraph{Various Task Types:}  To enhance the generalizability of our models, we gather datasets from a diverse range of task types. These include (see breakdown in \figureref{fig:task_type_breakdown}):
\begin{enumerate}
    \item \textbf{Pairwise Evaluation}: Tasks that involve comparing two responses at a time to determine a preference (e.g., \emph{``Which response, A or B, is more helpful?''}).
    \item \textbf{Pointwise Evaluation}: Tasks that involve evaluating specific attributes of individual responses independently (e.g., \emph{``Please rate the overall coherence of the response on a 5-point Likert scale.''}).
    \item \textbf{Classification}: Tasks that involve categorizing individual responses into predefined categories (e.g., \emph{``Does the model output follow the instructions? (Yes/No)''}). 
    \item \textbf{Open-ended Evaluation}: Tasks that require free-form, unrestricted answers (e.g., \emph{``Is the summary fully attributable to the source article? Provide a short explanation.''}).
\end{enumerate}

\paragraph{Various LLM Capabilities:} We choose datasets from literature that assess diverse LLM capabilities, including factuality, safety, reasoning, instruction-following, long-form generation, creativity, attribution, coding, etc. (see \sectionref{sec:capabilities}).

\subsection{LLM Capabilities Covered by FLAMe}
\label{sec:capabilities}

\begin{figure}[t]
\centering
\includegraphics[width=0.6\textwidth]{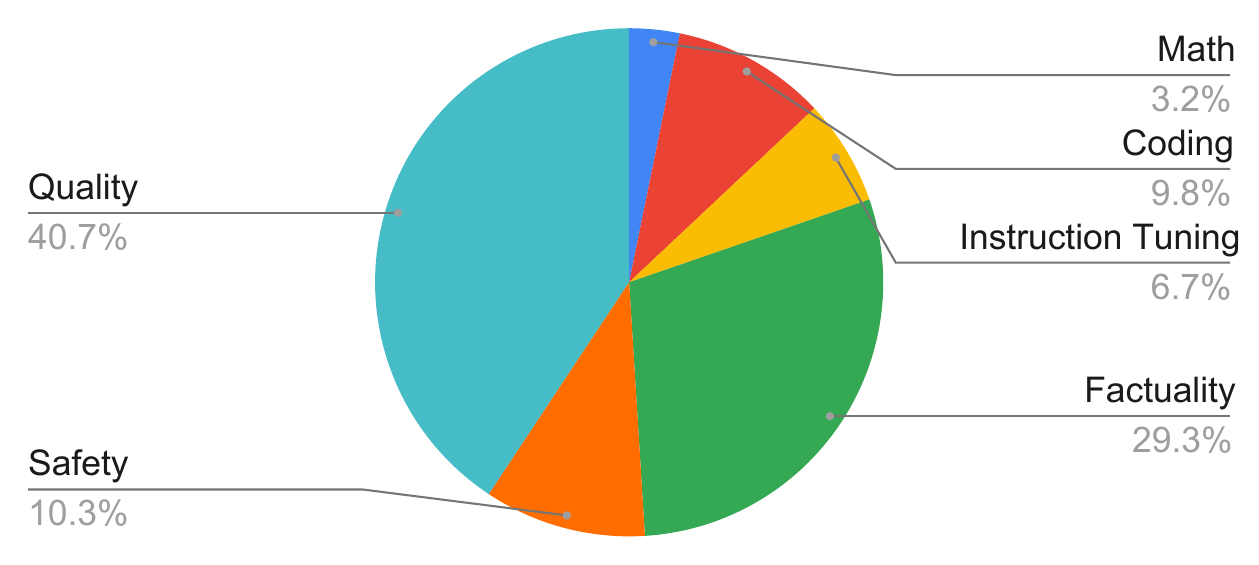}
\caption{A breakdown of our FLAMe data collection by LLM capability, with each slice representing the percentage of datapoints (out of 5.3M) for that specific LLM capability. We focus on the standard evaluation pillars regularly used in LLM evaluation: general response quality, factuality, safety, coding, and math. Additionally, we add some non-evaluation instruction tuning data (like LIMA) to help preserve the general-purpose instruction-following capabilities of FLAMe.}
\label{fig:llm_capability_breakdown}
\end{figure}

Following the principles outlined in \sectionref{sec:principles-training-data}, we curated a comprehensive data collection of 5.3M datapoints, spanning 102 training tasks (with an additional 53 tasks reserved for evaluation, as detailed in \sectionref{sec:evaluation-datasets}). ~\appendixref{appendix:training-data-list} contains the list of datasets used in our study. Our data collection encompasses key capabilities of contemporary LLMs, as outlined below (see breakdown in \figureref{fig:llm_capability_breakdown}).

\paragraph{General Response Quality:} To evaluate LLM response quality, we use various datasets that measure attributes like helpfulness, coherence, fluency, creativity, complexity, and verbosity. These include: Summary Comparisons (SummFeedback)~\citep{stiennon-etal-2020-learning}, LMSYS Chatbot Arena conversations~\citep{zheng-etal-2023-judging}, HH RLHF Helpfulness~\citep{bai-etal-2022-training}, WebGPT~\citep{nakano-etal-2021-webgpt}, SummEval~\citep{fabbri-etal-2021-summeval}, News Summary Evaluation~\citep{goyal-etal-2022-news}, SHP~\citep{ethayarajh-etal-2022-understanding}, BeaverTails Helpfulness~\citep{ji-etal-2023-beavertails}, SEAHORSE~\citep{clark-etal-2023-seahorse}, HelpSteer~\citep{wang-etal-2023-helpsteer}, etc. Additionally, to measure LLM instruction-following capabilities, we include datasets like GENIE~\citep{khashabi-etal-2022-genie}, InstruSum~\citep{liu-etal-2024-benchmarking}, and riSum~\citep{skopek-etal-2023-towards}.
\paragraph{Factuality/Attribution:} To address the increasing importance of measuring hallucinations in generated LLM responses, we incorporate several datasets
that evaluate the factual accuracy of responses and their grounding, measuring whether claims are supported by source documents. These include: XSum Hallucination~\citep{maynez-etal-2020-faithfulness}, QAGS~\citep{wang-etal-2020-asking}, WikiBio Hallucination~\citep{manakul-etal-2023-selfcheckgpt}, FRANK~\citep{pagnoni-etal-2021-understanding}, FactScore~\citep{min-etal-2023-factscore}, VitaminC~\citep{schuster-etal-2021-get}, HaluEval~\citep{li-etal-2023-halueval}, Q$^2$~\citep{honovich-etal-2021-q2}, FaithDial~\citep{dziri-etal-2022-faithdial}, DialFact~\citep{gupta-etal-2022-dialfact}, BEGIN~\citep{dziri-etal-2022-evaluating}, and MNLI~\citep{williams-etal-2018-broad}, etc.\footnote{We reformulate natural language inference as quality assessment because it naturally aligns with attribution.}

\vspace{-0.05in}
\paragraph{Mathematical Reasoning:} We construct datasets to help FLAMe differentiate between correct and incorrect solutions to mathematical problems. We leverage PRM800K~\citep{lightman-etal-2024-lets} and extract human vs incorrect LLM-generated solutions, as well as pairs of (\emph{correct, incorrect}) LLM-generated solutions. 
\vspace{-0.05in}
\paragraph{Coding:} In addition to natural language evaluation, we also train FLAMe to perform code evaluation. We utilize Code Contests~\citep{li-etal-2022-competition}, CommitPack~\citep{muennighoff-etal-2023-octopack}, and COFFEE~\citep{moon-etal-2023-coffee} to construct pairs of (\emph{correct, buggy}) programs in response to coding problems or GitHub issues. The model is trained to select the correct program or fix from each pair. Our training data covers popular programming languages, such as Python, JavaScript, Java, C++, Go, and Rust.
\vspace{-0.05in}
\paragraph{Safety:} Developing safe and harmless AI assistants for broad public use is increasingly important. To facilitate safety evaluation, we train FLAMe to identify more helpful and harmless responses. Our training data includes tasks from sources like HH RLHF Harmlessness~\citep{bai-etal-2022-training}, HH RLHF Red Teaming~\citep{ganguli-etal-2022-red}, BeaverTails QA-Classification and Harmlessness~\citep{ji-etal-2023-beavertails}.

\paragraph{Instruction Tuning:} Finally, to help preserve the instruction-following capabilities of our models, we incorporate instruction tuning data from datasets with human-written responses. These include: LIMA~\citep{zhou-etal-2023-lima}, PRM800K IF~\citep{lightman-etal-2024-lets},\footnote{We train the model to produce the ground truth solution for each problem.} and TULU-2~\citep{ivison-etal-2023-camels}.\footnote{We only use TULU-2 instruction tuning subsets with human-written responses, including FLAN, CoT, Open Assistant 1, Science literature, and Hardcoded (see Section 2 in~\citealp{ivison-etal-2023-camels} for details).}

\subsection{Unified Task Format}
\label{sec:curation}
\label{sec:unified-format}

After carefully selecting our training datasets (\sectionref{sec:principles-training-data}-\ref{sec:capabilities}), we process and standardize them into a unified text-to-text format. This preprocessing step typically takes about 3-4 hours per dataset and involves several key tasks:
\begin{enumerate}
    \item \textbf{Comprehensive Review and Author Consultations:} We carefully review the associated research and additionally consult with the original authors to clarify ambiguities or inconsistencies.
    \item \textbf{Data Collection:} We collect all relevant data files from the corresponding HuggingFace Datasets, TensorFlow Datasets, or GitHub repositories.
    \item \textbf{Data Extraction:} We identify and extract specific data fields containing quality assessments conducted by human annotators.
    \item \textbf{Task Definitions and Evaluation Instructions:} We meticulously create detailed task definitions and evaluation instructions for each quality assessment task, ensuring consistency and standardization. To maintain alignment with the original evaluation criteria, we adhere to any available instructions provided to the original human annotators. Our instructions help the model identify the input and output formats, as well as understand the specific aspects it should assess.
    \item \textbf{Text-to-Text Format Conversion:} Finally, we reformat all tasks as text-to-text tasks (see \figureref{fig:unified_task_format}). Task definitions, evaluation instructions, and desired output fields are listed under an \texttt{INSTRUCTIONS} block. Input field values and target field values are placed under \texttt{CONTEXT} and \texttt{EVALUATION} blocks, respectively. This flexible text-to-text format is easily adaptable to a wide range of quality assessment tasks.
\end{enumerate}

\section{Model}
We now leverage our large and diverse multitask mixture of quality assessment tasks to train general-purpose LLM autoraters, which can be prompted during inference to perform various tasks. We train three model variants: FLAMe, which is trained with examples-proportional mixture weights~\citep{raffel-etal-2020-exploring}; FLAMe-RM, which is initialized with FLAMe and slightly fine-tuned on a balanced mixture of four pairwise evaluation datasets, spanning chat, reasoning, and safety (\sectionref{sec:flame-rm}); and FLAMe-Opt-RM, which is trained with reward modeling optimized mixture weights, determined using a tail-patch fine-tuning strategy (\sectionref{sec:tail-patch-algo}).
\subsection{Training General-purpose LLM Autoraters (FLAMe)}
\label{sec:flame-general}
We start with a baseline training approach by using supervised multitask training to train an instruction-tuned PaLM-2-24B model on our multitask mixture for a fixed number of 30K training steps. We employ examples-proportional mixture weights, capped at a maximum of $2^{16}$ per task to avoid oversampling large datasets. Our resulting FLAMe model significantly improves generalization to a diverse array of held-out tasks, outperforming models like GPT-4, Claude-3, and Llama-3 on many tasks (see \figureref{fig:rewardbench_performance} and \tableref{tab:main_results_table}). These findings support our hypothesis that large-scale multitask instruction tuning effectively equips the model with general-purpose quality assessment capabilities. However, we find that this approach is not optimal for specialized downstream applications like reward modeling evaluation, which motivates our approaches targeting specific downstream distributions (\sectionref{sec:flame-rm} and \sectionref{sec:tail-patch-algo}).
\subsection{Fine-tuning FLAMe for Reward Modeling Evaluation (FLAMe-RM)}
\label{sec:flame-rm}
Motivated by our findings with FLAMe, we delve deeper into the potential of FLAMe as a powerful starting point for further fine-tuning on specific downstream applications. We focus on reward modeling evaluation as a case study. We create FLAMe-RM by fine-tuning FLAMe on a mixture of four pairwise evaluation datasets, equally mixed, spanning chat, reasoning, and safety. These include: HelpSteer~\citep{wang-etal-2023-helpsteer}, PRM800K~\citep{lightman-etal-2024-lets}, CommitPack~\citep{muennighoff-etal-2023-octopack}, and HH-RLHF Harmlessness~\citep{bai-etal-2022-training}.
Since FLAMe is already trained on these datasets, we only fine-tune it for 50 steps. The resulting FLAMe-RM model significantly improves the original FLAMe's RewardBench overall score from 86.0\% to 87.8\% accuracy. Remarkably, FLAMe-RM-24B is the top-performing generative model trained exclusively on permissively licensed data, surpassing both GPT-4-0125 (85.9\%) and GPT-4o (84.7\%); see \figureref{fig:rewardbench_performance} and \tableref{tab:main_results_table}.

\subsection{Optimizing FLAMe Multitask Mixture for Reward Modeling Evaluation (FLAME-Opt-RM)}
\label{sec:tail-patch-algo}
\begin{figure}[t]
\centering
\includegraphics[width=0.48\textwidth]{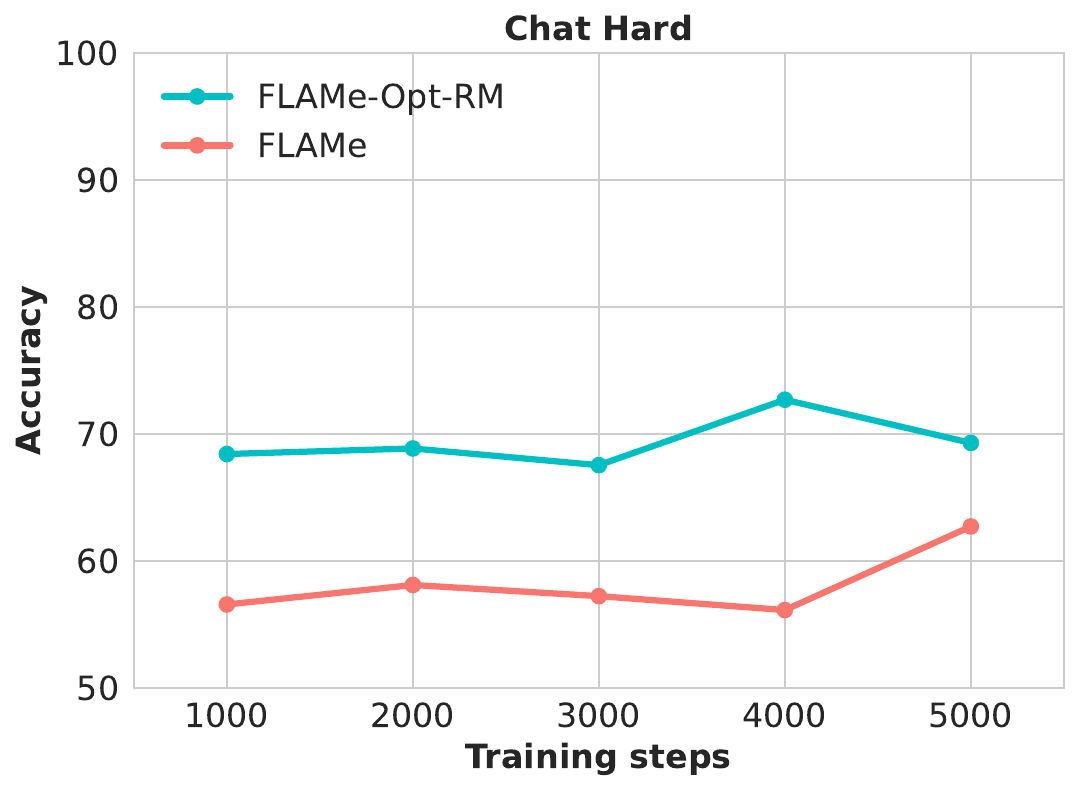}
\includegraphics[width=0.48\textwidth]{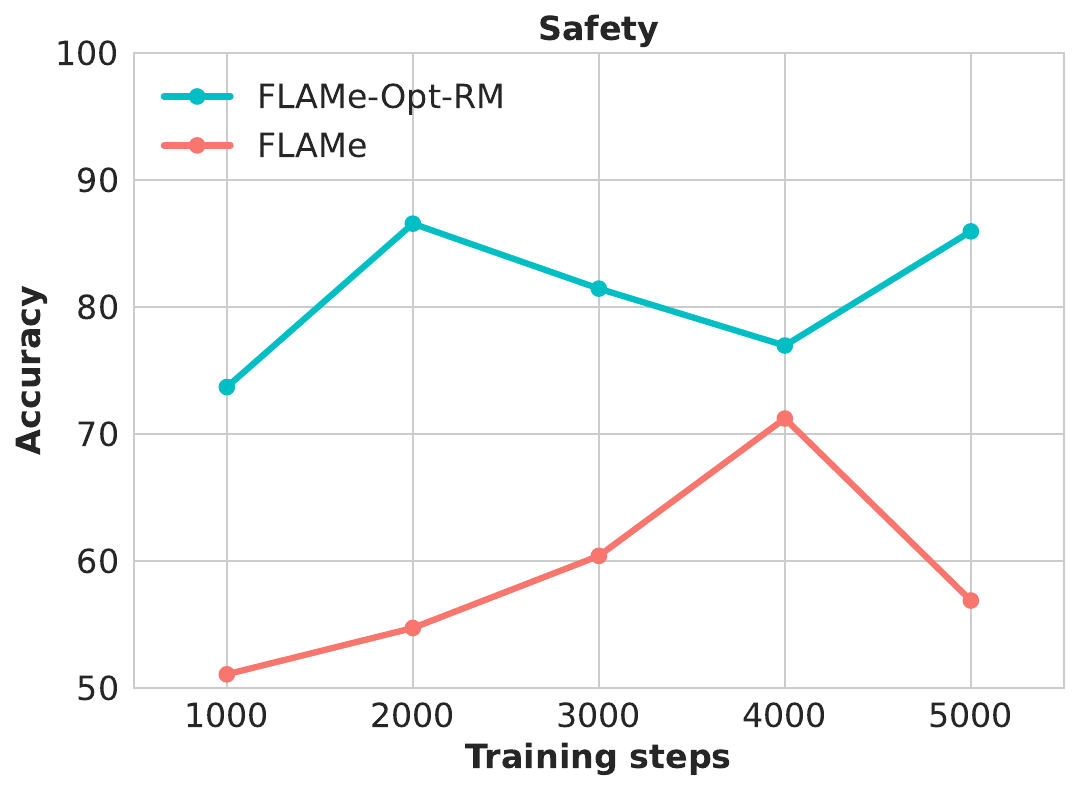}
\caption{A comparison of FLAMe-Opt-RM and FLAMe in early training stages (first 5000 steps) based on RewardBench Chat Hard and Safety performance. FLAMe-Opt-RM, with optimized mixture weights, achieves significantly higher Chat Hard and Safety scores faster than FLAMe. For reference, FLAMe achieves Chat Hard and Safety scores of 66.2 and 88.5, respectively, at 30K training steps.}
\label{fig:chat_hard_performance}
\end{figure}
While our vanilla FLAMe mixture with examples-proportional mixing performs well across many tasks, it requires extensive training to attain strong performance on certain specialized downstream applications, for example, RewardBench (see \figureref{fig:chat_hard_performance}). We attribute this to suboptimal mixture weights that undersample beneficial tasks during training. To address this, we introduce a novel tail-patch ablation strategy that analyzes the impact of each dataset on targeted distributions. This allows us to find the optimal proportions of individual datasets in our multitask mixture, efficiently optimizing all mixing weight hyperparameters at once. By fine-tuning the initial instruction-tuned PaLM-2-24B checkpoint on this optimized mixture for only 5000 steps, we achieve competitive RewardBench performance (87.0\%) with our baseline FLAMe approach (86.0\%) while using approximately 25$\times$ less training datapoints. 

Here, we directly optimized our multitask mixture based on RewardBench performance changes due to its lack of a development set. In early experiments, we observed weak correlations between RewardBench performance and performance on our other held-out tasks across model variants, preventing us from creating a reliable ``proxy'' development set. We emphasize that our goal here is not to achieve state-of-the-art RewardBench results but instead to demonstrate how our multitask mixture can be optimized for targeted distributions. We found that longer training and/or additional fine-tuning, as is done for FLAMe-RM, further improved our RewardBench performance, though we did not submit these FLAMe-Opt-RM results to the official RewardBench leaderboard. Furthermore, FLAMe-Opt-RM's robust performance across other held-out tasks (see \tableref{tab:main_results_table}) indicates that we have not overfitted to RewardBench, affirming the broad applicability of FLAMe-Opt-RM across diverse tasks.

\paragraph{Tail-patch Ablations to Determine Beneficial Tasks:} Setting the right mixing weight for each individual training task in our multitask mixture is non-trivial due to the large number of tasks. Instead, we examine the impact of each task on targeted distributions, and then use this information for weight assignment. First, we select a checkpoint that is partially trained on our vanilla mixture, showing decent but not optimal performance across RewardBench categories.\footnote{We hypothesize that using a partially trained checkpoint, rather than the initial one, is better for tail-patch ablations, since the model has already been exposed to multitask data and is familiar with its overall distribution.} Then, we perform a brief fine-tuning stage (``tail-patch'') \emph{exclusively} on each individual training task, limited to 3000 training steps. We posit that training on a beneficial task would bridge the gap between fair and optimal performance. We note that this is a one-time procedure per downstream application and can be done with smaller models to further reduce computational costs.
\vspace{0.05in}

\noindent \textbf{A Re-weighted Mixture Based on Tail-patch Ablations}: After training a tail-patch on each task, we rate how helpful each training task is to each category of RewardBench using one of four ratings: \emph{Helpful} (+2, performance significantly improves and remains stable), \emph{Somewhat helpful} (+1, performance slightly improves), \emph{No clear effect} (0, performance is nearly unchanged), \emph{Harmful} (-1, performance is significantly worse). We then organize tasks into seven bundles: \emph{Generally helpful} (tasks with the highest total ratings, $\geq 5$ in our study), \emph{Category-specific}, one for each of the five RewardBench categories (most beneficial tasks for a specific category where performance crosses a threshold $\tau$),\footnote{We separate Math and Coding for the Reasoning category, and use thresholds of $\tau=95\%, 66\%, 99.8\%, 84\%, 85\%$ for Chat, Chat Hard, Math, Coding, and Safety, respectively.} and \textit{Others} for the remaining tasks.

We assign a fixed mixing weight for each bundle: $w_{general}$=100K for \emph{Generally helpful}, $w_{specific}$=30K for each \emph{Category-specific} bundle, and $w_{others}$=3K for \textit{Others}. A task can belong to more than one bundle; in this case, its final weight is the sum of the mixture weights from all the bundles it belongs to. For example, if a task is generally helpful and specifically beneficial for both Chat Hard and Safety, it contributes $w_t = w_{general} + 2 \times w_{specific}$ to our final mixture. An exception to this rule: we prioritize the top two most helpful tasks in three categories with suboptimal performance—--Chat Hard, Coding, and Safety--—each with a fixed weight of $w_{top\_specific}$=200K. These weight values were initially set based on our intuition and were not extensively tuned. FLAME-Opt-RM is initialized with the initial instruction-tuned PaLM-2-24B and then fine-tuned using our re-weighted multitask mixture.
\subsection{Training Details}
\label{sec:training}

We initialize both FLAMe and FLAMe-Opt-RM with the PaLM-2-24B model~\citep{anil-etal-2023-palm}, instruction-tuned on the Flan collection~\citep{chung-etal-2024-scaling,longpre-etal-2023-the}, and train for 30K and 5K steps, respectively. FLAMe is then further fine-tuned for 50 steps to create FLAMe-RM. Our models are trained using T5X~\citep{roberts-etal-2023-scaling}  with the Adam optimizer~\citep{kingma-ba-2015-adam}, a learning rate of 0.0001, and a dropout rate of 0.05. FLAMe is trained on 256 Cloud TPU chips with a batch size of 32, whereas FLAMe-RM and  FLAMe-Opt-RM use 128 Cloud TPU chips with a batch size of 8.\footnote{\url{cloud.google.com/tpu/docs/v5e-training}, \url{https://cloud.google.com/tpu/docs/v3}}

\section{Experiments}
\label{sec:experiments}

Having discussed our FLAMe variants and their implementations in~\sectionref{sec:flame-modeling}, we now present our main experiments. We compare FLAMe to several popular LLM-as-a-Judge autoraters (\sectionref{sec:baselines}) using an evaluation suite that includes 12 autorater benchmarks: 1 held-in and 11 held-out, covering a total of 53 quality assessment tasks (\sectionref{sec:evaluation-datasets}). Overall, we find that our FLAMe variants, trained exclusively on permissively licensed data, outperform LLMs trained on proprietary data like GPT-4 and Claude-3 on 8 out of 12 benchmarks (\sectionref{sec:main_results}).

\subsection{Evaluation Datasets}
\label{sec:evaluation-datasets}

Our goal is to measure general quality assessment capabilities of FLAMe. As such, we evaluate our models using a diverse set of held-in and held-out tasks. We cast each task into our unified text-to-text format (\sectionref{sec:unified-format}) and prompt our models to perform the task. For benchmarks with multiple categories (e.g., RewardBench and LLM-AggreFact), we use the same prompt instructions across categories. To reduce model API costs, we randomly sample 256 examples per evaluation task,\footnote{For tasks with fewer than 256 examples, we use the full evaluation set.} except for RewardBench, where we report results on the full evaluation sets.

\subsubsection{Held-in Evaluations}
\paragraph{HelpSteer~\citep{wang-etal-2023-helpsteer}:} We assess FLAMe's performance on rating helpfulness, correctness, coherence, complexity, and verbosity, using the HelpSteer validation set.

\subsubsection{Held-out Evaluations}

\paragraph{RewardBench~\citep{lambert-etal-2024-rewardbench}:} RewardBench is a widely used benchmark for assessing reward models, focusing on their capabilities and safety. It involves pairwise preference tasks where reward models choose the better response between two options based on a given prompt. RewardBench encompasses four main categories aimed at evaluating specific desired capabilities in LLMs: Chat, Chat Hard, Reasoning (Math + Coding), and Safety. The benchmark incorporates 23 individual datasets.\footnote{We excluded the ``Prior sets'' of RewardBench because three out of the four datasets were used in training FLAMe.}

\paragraph{LLM-AggreFact~\citep{tang-etal-2024-minicheck}:} LLM-AggreFact is a benchmark for measuring the grounding capabilities of autoraters. Given a reference document and a claim, the autorater determines if the claim is fully supported by the document. This holistic benchmark combines 10 attribution datasets used in recent studies on LLM factuality.

\paragraph{Other Evaluation Benchmarks:} In addition to RewardBench and LLM-AggreFact, we evaluate FLAMe on a diverse array of other held-out pairwise comparison and pointwise evaluation benchmarks, including: Summary Comparisons (SummFeedback)~\citep{stiennon-etal-2020-learning};\footnote{During training, we used only pairwise ratings from the dataset and reserved pointwise ratings for evaluation.} Helpful, Honest, and Harmless Alignment (HHH)~\citep{askell-etal-2021-general}; AlpacaFarm~\citep{dubois-etal-2023-alpacafarm}; Paraphrase Evaluation (Dipper)~\citep{krishna-etal-2023-paraphrasing}; Sequence Continuation Preference  (RankGen)~\citep{krishna-etal-2022-rankgen}; Poem Preference (CoPoet)~\citep{chakrabarty-etal-2022-help};  Literary Translation Comparisons (LitTrans)~\citep{karpinska-iyyer-2023-large}; Long-form QA Evaluation (LFQAEval)~\citep{xu-etal-2023-critical}; and Text Continuation Preference (ContrSearch)~\citep{su-etal-2022-empirical}. None of the tasks in these benchmarks were included in our training data.

\subsection{Evaluated Models}
\label{sec:baselines}
We evaluate several popular LLM-as-a-Judge models as baselines, including: Llama-3-70B-Instruct~\citep{meta-2024-introducing}, Mixtral 8$\times$7B~\citep{jiang-etal-2024-mixtral}, Claude-3-Opus~\citep{anthropic-2024-introducing}, GPT-3.5-turbo-0125~\citep{openai-2024-chatgpt}, GPT-4-0125~\citep{openai-2024-gpt4-turbo}, and OpenAI's current flagship model GPT-4o~\citep{openai-2024-gpt4o}.\footnote{For fair comparison, we use the same FLAMe prompt instructions when evaluating LLM-as-a-Judge baselines. For better reproducibility, we set the temperature to 0 and generate up to 1024 tokens across all models.} We also compare our results with several models on the official RewardBench leaderboard, notably Gemini-1.5-Pro~\citep{reid-etal-2024-gemini}, Prometheus-2-8$\times$7B~\citep{kim-etal-2024-prometheus-2}, and NVIDIA's Nemotron-4-340B-Reward and Llama-3-70B-SteerLM-RM~\citep{wang-etal-2024-helpsteer2}.

We evaluate all our three FLAMe variants: FLAMe, FLAMe-RM, and FLAMe-Opt-RM, as described in \sectionref{sec:flame-general}-\sectionref{sec:tail-patch-algo}. Additionally, we include the initial instruction-tuned PaLM-2-24B checkpoint, which has not been trained on our FLAMe data, to separate the impact of instruction tuning and FLAMe training.

\subsection{Main Results}
\label{sec:main_results}

\begin{table*}[t]
\centering
\begin{adjustbox}{max width=\textwidth}
\begin{tabular}{l c c c c c c c c c c c c}
\toprule
Model & Reward &  LLM   & Summ & Alpaca & Rank & Co & Contr & HHH & Dipper & Lit & LFQA & Help \\[-0.75mm]
& Bench & AggreFact  & Feedback & Farm & Gen & Poet & Search & & & Trans & Eval & Steer \\

\midrule
Llama-3-70B-Instruct & 76.1 & 76.1 & 50.8 & 53.9 & 65.6 & 53.6 & 53.1 & 91.9 & 42.8 & 60.5 & 71.1 & 39.7 \\
Mixtral-8$\times$7B & 77.8 & 73.8 & 43.8 & 55.1 & 63.3 & 52.9 & 56.6 & 90.0  & 42.2 & 61.7 & 71.5 & 34.0 \\
GPT-3.5-turbo-0125 & 64.5 & 70.0 & 15.6 & 55.5 & 58.2 & 49.0 & 57.5 & 85.5 & 45.0 & 54.3 & 69.9 & 32.0 \\
Claude-3-Opus & 80.7 & 79.2 & 31.6 & 49.6 & 55.1 & 49.0 & 45.1 & \bf 94.6 & \bf 50.6 & 71.1  & 71.1 & 41.3 \\
GPT-4-0125 & 85.9 & 80.6 & 46.5 & 49.6 & 62.5 & 56.9 & 55.8 & \bf 94.6 & 45.0 & 67.6 & \bf 77.0 & 37.9 \\
GPT-4o & 84.7 & 80.2 & 30.9 & 50.4 & 66.0 & 55.6 & 57.5 & 92.3 & 45.6 & \bf 72.7 & 75.0 & 40.1 \\

\midrule
\multicolumn{5}{l}{\emph{our models}} \\[2mm]
PaLM-2-24B & 62.9 & 54.8 & 13.3 & 52.3 & 58.2 & 54.2 & 46.0 & 85.5 & 48.3 & 62.5 & 70.3 & 20.0 \\
FLAMe-24B & 86.0 & \textbf{\textcolor{blue}{81.1}} & 48.0 & \textbf{\textcolor{blue}{58.2}} & 62.1 & 53.6 & \textbf{\textcolor{blue}{69.9}} & 91.4 & 48.3 & 67.2 & 74.2 & \textbf{\textcolor{blue}{48.4}} \\
FLAMe-RM-24B & \textbf{\textcolor{blue}{87.8}} & 80.8 & \textbf{\textcolor{blue}{53.1}} & 57.8 & 65.2 & \textbf{\textcolor{blue}{57.5}} & 57.5 & 91.0 & 47.8 & 67.6 & 72.7 & 46.6 \\
FLAMe-Opt-RM-24B & 87.0 & 80.2 & 52.3 & 53.1 & \textbf{\textcolor{blue}{69.5}} & 52.9 & 48.7 & 89.1 & 48.3 & 69.5 & 69.5 & 35.9 \\
\bottomrule
\end{tabular}
\end{adjustbox}
\caption{Performance of FLAMe compared to LLM-as-a-Judge baselines across a wide variety of autorater evaluation benchmarks. Overall, FLAMe variants outperform all popular proprietary LLM-as-a-Judge models on 8 out of 12 benchmarks,including RewardBench and LLM-AggreFact. See \sectionref{sec:evaluation-datasets} for the sources of our evaluation benchmarks.}%
\label{tab:main_results_table}
\end{table*}
\begin{table}[t]
\small
\centering
\begin{tabular}{l c c c c c }
\toprule
Model & Average & Chat & Chat Hard & Safety & Reasoning \\
\midrule
\multicolumn{6}{l}{\emph{custom classifiers on the official RewardBench leaderboard}} \\[2mm]
Nemotron-4-340B-Reward & \textbf{92.2} & 95.8 & \textbf{87.1} & 92.2 & 93.6 \\
Cohere May 2024 & 89.5 & \textbf{96.4} & 71.3 & 92.7 & \textbf{97.7} \\
Llama3-70B-SteerLM-RM & 89.0 & 91.3 & 80.3 & \textbf{93.7} & 90.6 \\
\midrule
\multicolumn{6}{l}{\emph{generative models on the official RewardBench leaderboard}} \\[2mm]
GPT-3.5-turbo-0125 & 64.5 & 92.2 & 44.5 & 62.3 & 59.1 \\
Prometheus-2-8$\times$7B & 75.3 & 93.0 & 47.1 & 83.5 & 77.4 \\
Llama-3-70B-Instruct & 76.0 &  \bf 97.6 & 58.9 & 69.2 & 78.5 \\
Mixtral-8$\times$7B & 77.8 & 95.0 & 64.0 & 73.4 & 78.7 \\
Claude-3-Opus & 80.7 & 94.7 & 60.3 &  89.1 & 78.7 \\
Gemini-1.5-Flash & 82.1 & 92.2 & 63.5 & 87.7 & 85.1 \\
GPT-4o &  84.7 & 96.6 & 70.4 & 86.7 & 84.9 \\
GPT-4-0125 & 85.9 & 95.3 & 74.3 & 87.2 & 86.9 \\
Gemini-1.5-Pro & \bf 88.1 & 92.3 & \bf  80.6 & 87.5 & 92.0 \\[2mm]

\multicolumn{5}{l}{\emph{our generative autorater models}} \\[2mm]

PaLM-2-24B & 62.9 & 89.9 & 61.2 & 55.3 & 45.2 \\
FLAMe-24B & 86.0 & 94.7 & 66.2 & 88.5 & \textbf{\textcolor{blue}{94.7}} \\
FLAMe-RM-24B & 87.8 & 92.2 & 75.7 & \textbf{\textcolor{blue}{89.6}} & 93.8 \\
FLAME-Opt-RM-24B & 87.0 & 92.2 & 77.0 & 86.2 & 92.5 \\
\bottomrule
\end{tabular}
\caption{A comparison of FLAMe with other generative models on the official RewardBench leaderboard. FLAMe-RM-24B achieves the best overall performance (87.8\%) among generative models trained solely on permissively licensed data.}
\label{tab:rewardbench_table}
\end{table}
\begin{table}[h!]
\small
\centering
\begin{tabular}{l c c c c c }
\toprule
Model & Overall  & LLM-FactVerify & Wiki-FactVerify & Summarization & Long-form QA \\
\midrule
GPT-3.5-turbo-0125 & 70.0 & 80.1 & 71.1 & 64.6 & 65.4 \\
Mixtral-8$\times$7B & 73.8 & 73.8 & 50.8 & 78.1 & 76.6 \\
Llama-3-70B-Instruct & 76.1 & 75.3 & 58.4 & 80.3 & \bf 77.7 \\
Claude-3-Opus & 79.2 & 78.6 & 70.6 & 83.8 & 75.0 \\
GPT-4o & 80.2 & 79.6 & 71.6 & 85.0 & 76.0 \\
GPT-4-0125 & 80.6 & 79.6 & 71.6 & 85.3 & 77.3 \\
\midrule
\multicolumn{5}{l}{\emph{our models}} \\
PaLM-2-24B & 54.8 & 34.4 & 28.9 & 68.2 & 71.7 \\
FLAMe-24B & \textbf{\textcolor{blue}{81.1}} & 82.3 & 77.7 & 85.3 & 72.7 \\
FLAMe-RM-24B & 80.8 & \textbf{\textcolor{blue}{82.6}} & 77.2 & \textbf{\textcolor{blue}{85.4}} & 70.9 \\
FLAMe-Opt-RM-24B & 80.2 & 77.6 & \textbf{\textcolor{blue}{81.2}} & 84.7 & 74.8 \\
\bottomrule
\end{tabular}
\caption{LLM-AggreFact performance across four common use-cases: LLM-FactVerify (ClaimVerify + FactCheck + Reveal), Wiki-FactVerify (WiCE), Summarization (AggreFact + TofuEval), and Long-form QA (ExpertQA + LFQA). FLAMe variants outperform all tested LLM-as-a-Judge models in three out of the four use-cases. FLAMe-24B achieves the highest overall performance of 81.1, while the next-best model GPT-4-0125 scores 80.6.}
\label{tab:llmaggfact_table}
\end{table}

\noindent \tableref{tab:main_results_table} shows our main results across all evaluation benchmarks. RewardBench and LLM-AggreFact results are shown in \tableref{tab:rewardbench_table} and \tableref{tab:llmaggfact_table}, respectively. Below, we first provide an overview of these results before
analyzing them in more detail:

\paragraph{FLAMe Variants Outperform all LLM-as-a-Judge baselines on 8 out of 12 benchmarks:} Our results in \tableref{tab:main_results_table} suggest that FLAMe variants, despite being trained solely on permissively licensed datasets, perform strongly across various evaluation benchmarks. Remarkably, our models outperform all state-of-the-art LLM-as-a-Judge models trained on proprietary data on 8 out of 12 benchmarks. FLAMe variants exceed the next-best model by a large margin on several held-out benchmarks, including: ContrSearch (69.9 vs 57.5 for GPT-4o/GPT-3.5-turbo-0125), RankGen (69.5 vs 66.0 for GPT-4o), AlpacaFarm (58.2 vs 55.5 for GPT-3.5-turbo-0125), SummFeedback (53.1 vs 50.8 for Llama-3-70B-Instruct), and RewardBench (87.8 vs 85.9 for GPT-4-0125). Unsurprisingly, our models also obtain the best held-in performance on HelpSteer (48.4 vs. 41.3 for Claude-3-Opus). On the other hand, FLAMe variants lag behind proprietary models on several benchmarks, including HHH (91.4 vs 94.6 for GPT-4-0125/Claude-3-Opus), LitTrans (69.5 vs 72.7 for GPT-4o), and LFQAEva (74.2 vs 77.0 for GPT-4-0125), suggesting that these proprietary models may have been optimized for these capabilities. Interestingly, GPT-4-0125 outperforms GPT-4o on 6 out of 12 benchmarks, including RewardBench, despite GPT-4o achieving a higher rank on the official LMSYS leaderboard~\citep{chiang-etal-2024-chatbot}. Finally, FLAMe provides significant gains over the initial instruction-tuned PaLM-2-24B across almost all benchmarks, highlighting the benefits of FLAMe training. Overall, our results demonstrate FLAMe's robust generalization to held-out tasks, showcasing its effectiveness as a versatile LLM autorater.

\paragraph{FLAMe Variants Are Among The Most Powerful Generative Models on RewardBench:} Our results in \tableref{tab:rewardbench_table} indicate that FLAMe variants are among the top-performing generative models on the official RewardBench leaderboard, achieving strong performance across all categories: Chat, Chard Hard, Safety, and Reasoning. Notably, FLAMe-RM-24B achieves an overall score of 87.8\%, the best performance among generative models trained solely on permissively licensed data, surpassing both GPT-4-0125 (85.9) and GPT-4o (84.7). 
As of July 15, 2024, FLAMe-RM-24B ranks second among generative models (below Gemini-1.5-Pro) and sixth among all models (spanning various model types such as custom classifier, generative, sequence classifier, and DPO) on the official RewardBench leaderboard.\footnote{\url{https://huggingface.co/spaces/allenai/reward-bench}.} While RewardBench is a widely used benchmark for evaluating reward models, we identified issues with length and token bias during our evaluations. We provide an analysis of bias in RewardBench in \appendixref{appendix:rb-analysis}.

\paragraph{FLAMe Attains the Best Performance on LLM-AggreFact:} Finally, \tableref{tab:llmaggfact_table} presents our attribution results on LLM-AggreFact~\citep{tang-etal-2024-minicheck}, categorized into four common use-cases: 1) LLM-FactVerify: fact verification of LLM-generated responses, 2) Wiki-FactVerify: evaluating correctness of Wikipedia claims, 3) Summarization: assessing faithfulness of summaries, and 4) Long-form QA: evaluating long-form answers to questions. FLAMe variants outperform all other models in three out of the four categories (LLM-FactVerify, Wiki-FactVerify, and Summarization). FLAMe-24B achieves the highest overall performance of 81.1, while the next-best baseline model GPT-4-0125 obtains a score of 80.6. In long-form QA attribution evaluation, our best model FLAMe-Opt-RM underperforms compared to GPT-4-0125 (74.8 vs 77.3), aligning with our findings in \tableref{tab:main_results_table}.

\section{Further Analysis of FLAMe}
\label{sec:applications}
In this section, we provide an analysis to elucidate some interesting aspects of our models. We depart from the usual focus on analyzing the effect of factors like model size, data size, and data quality in multitask learning, which have been extensively studied in recent work on multitask learning and instruction tuning~\citep{raffel-etal-2020-exploring,longpre-etal-2023-the}. Instead, we explore potential biases inherent in our LLM autoraters. Additionally, we demonstrate the potential utility of FLAMe for AI development, such as sampling high-quality responses.

\subsection{Autorater Bias Analysis}
\label{sec:autorater-bias}

\begin{table*}[t!]
\scriptsize
\begin{center}
\begin{tabular}{ lccccccc } 
 \toprule
Autorater & Avg. ($\downarrow$) & Order ($\downarrow$) & Compassion ($\downarrow$) & Length ($\downarrow$) & Egocentric ($\downarrow$) & Bandwagon ($\downarrow$) & Attention ($\downarrow$) \\
\midrule
Random & 0.30 & 0.50 & 0.50 & 0.00 & 0.25 & 0.25 & 0.25  \\
\midrule
\multicolumn{4}{l}{\emph{baselines reported in~\citet{koo-etal-2023-benchmarking}}} \vspace{0.01in}\\
Falcon-40B & 0.31 & 0.77 & 0.27 & 0.09  & \bf 0.05 & 0.28 & 0.40 \\
Cohere-54B & 0.41 & 0.50 & 0.65 & 0.10  & 0.27 & 0.82 & 0.14 \\
Llama-2-70B & 0.19 & 0.61 & 0.26 & 0.12  & 0.06 & 0.04 & 0.03 \\
InstructGPT & 0.45 & 0.38 & 0.48 & 0.16  & 0.28 & 0.85 & 0.54 \\
ChatGPT & 0.45 & 0.41 & 0.66 & 0.13  & 0.58 & 0.86 & 0.06 \\
GPT-4 & 0.31 & 0.23 & 0.79 & 0.06  & 0.78 & \bf 0.00 & \bf 0.00 \\
 \midrule
 \multicolumn{4}{l}{\emph{our models}} \vspace{0.01in}\\
 FLAMe-24B & \textbf{\textcolor{blue}{0.13}} & \textbf{\textcolor{blue}{0.08}} & 0.09 & 0.03 & 0.38 & 0.18 & \textbf{\textcolor{blue}{0.00}} \\
 FLAMe-RM-24B & \textbf{\textcolor{blue}{0.13}} & 0.11 & \textbf{\textcolor{blue}{0.08}} & 0.02 & 0.40 & 0.17 & \textbf{\textcolor{blue}{0.00}} \\
 FLAMe-Opt-RM-24B & 0.15 & 0.15 & 0.14 & \textbf{\textcolor{blue}{0.00}} & 0.41 & 0.17 & \textbf{\textcolor{blue}{0.00}} \\
\bottomrule
\end{tabular}
\end{center}
\caption{Autorater bias analysis on the CoBBLEr bias benchmark from~\citet{koo-etal-2023-benchmarking}. \textbf{Lower values indicate better or less biased autoraters} across all columns. Overall, we find that FLAMe variants exhibit significantly less bias compared to popular LLM-as-a-Judge autoraters like GPT-4. Compared to Table 2 in~\citet{koo-etal-2023-benchmarking}, we combine first/last numbers for Order/Compassion, report $|\text{bias} - 0.5|$ for Length, and exclusively report the order variant in Egocentric.}
\label{tab:cobbler-results}
\end{table*}

A common criticism of LLM-as-a-Judge autoraters involves their bias towards certain judgments~\citep{liu-etal-2023-g,panickssery-etal-2024-llm,liu-etal-2023-llms,bai-etal-2023-benchmarking}. In this section, we evaluate FLAMe variants on the CoBBLEr autorater bias benchmark~\citep{koo-etal-2023-benchmarking}. We find that our models are significantly less biased than other popular LLM-as-a-Judge autoraters.

CoBBLEr measures six types of biases in LLM autoraters: 

\begin{enumerate}
    \item \textbf{Order:} Does the autorater have a preference towards the response position?
    \item \textbf{Compassion:} Does the autorater's judgment change when the response-generating LLM's actual name, such as ``GPT-4'', is used instead of aliases like ``Model A''?
    \item \textbf{Length:} Does the autorater have a preference for longer or shorter outputs?
    \item \textbf{Egocentric:} Does the autorater have a preference for outputs generated by itself?
    \item \textbf{Bandwagon:} Does the autorater get swayed by sentences like \emph{``90\% people prefer response A''}?
    \item \textbf{Attention:} Does the autorater get distracted by irrelevant context, such as \emph{``Response A is about cats.''}?
\end{enumerate}

We leverage the original (\emph{prompt,response}) pairs from~\citet{koo-etal-2023-benchmarking} and reformat them into our unified FLAMe format (\figureref{fig:unified_task_format}). We compare FLAMe variants to other LLM-as-a-Judge autoraters reported in~\citet{koo-etal-2023-benchmarking}, including GPT-4.

Our results are shown in \tableref{tab:cobbler-results}. We find that FLAMe variants exhibit significantly lower bias compared to GPT-4 and other autoraters, with an average bias of 0.13 vs 0.31 for GPT-4 (lower is better). FLAMe yields significantly better or on-par performance compared to GPT-4 across all six bias categories. These results demonstrate FLAMe's effectiveness as a robust and reliable autorater.

\begin{table}[t]
\small
\centering
\begin{tabular}{l c c c }
\toprule
Ranker & CodeGen-16B & davinci002 & InCoder-6B  \\
\midrule
\multicolumn{4}{l}{\emph{10 code samples re-ranked in round-robin fashion}} \\[2mm]
None & 21.2 & 17.6 & 14.6 \\
 FLAMe-24B & \textbf{\textcolor{blue}{31.1}} & 22.6 & \textbf{\textcolor{blue}{22.0}} \\
 FLAMe-RM-24B  & 29.9 & \textbf{\textcolor{blue}{23.2}} & 21.3  \\
 FLAME-Opt-RM-24B  & 29.3 & 18.3 & 16.5 \\
\midrule
Oracle & 46.9 & 63.4 & 29.3 \\
\bottomrule
\end{tabular}

\caption{Pass@1 performance on the HumanEval coding benchmark~\citep{chen-etal-2021-evaluating}. Re-ranking code samples with FLAMe variants significantly improves performance across models.}
\label{tab:humaneval-reranking}

\end{table}

\subsection{Using FLAMe to Re-rank Decoded Outputs}
\label{sec:reranker-results}
Finally, we explore the application of our LLM autoraters in selecting optimal outputs from multiple responses, a method known as \emph{``Best-of-N''} sampling~\citep{nakano-etal-2021-webgpt,krishna-etal-2022-rankgen}. Using FLAMe for re-ranking, we assess its impact on code generation performance with the HumanEval Python programming benchmark~\citep{chen-etal-2021-evaluating}. We conduct experiments by re-ranking 10 code samples generated by three models: OpenAI's davinci-002, InCoder-6B~\citep{fried-etal-2023-incoder}, and CodeGen-16B~\citep{nijkamp-etal-2023-codegen} using a round-robin competition, and then measuring performance with the top-ranked code sample.\footnote{We use relatively weak LLMs from~\citet{chen-etal-2023-codet} for two main reasons: (1) to assess the potential benefits of re-ranking with FLAMe, and (2) HumanEval has been extensively used to develop newer LLMs.} Results in \tableref{tab:humaneval-reranking} show that FLAMe provides significant gains in pass@1 accuracy across all three models. Notably, FLAMe improves CodeGen-16B's pass@1 from 21.2 to 31.1, closing nearly 40\% of the gap to the Oracle ranker (46.9).

\section{Conclusion}
\label{sec:conclusion}
We introduce FLAMe, a family of foundational autorater models that can perform various quality assessment tasks. FLAMe is trained on a large and diverse collection of curated and standardized human evaluations derived exclusively from permissively licensed datasets. We demonstrate FLAMe's strong zero-shot generalization abilities, outperforming models trained on proprietary data like GPT-4 and Claude-3 on many held-out tasks. FLAMe can also effectively serve as a powerful starting point for further downstream fine-tuning. Our FLAMe-RM variant, which is fine-tuned for reward modeling evaluation, is among the top-performing generative models on RewardBench, despite being trained solely on permissively licensed data, outperforming both GPT-4-0125  and GPT-4o. Additionally, we present a more computationally efficient approach using a novel tail-patch fine-tuning strategy to optimize our FLAMe multitask mixture for targeted distributions, offering competitive performance with significantly less compute. Our FLAMe variants outperform popular proprietary LLM-as-a-Judge models across 8 out of 12 autorater evaluation benchmarks, covering 53 quality assessment tasks, including RewardBench and LLM-AggreFact. Finally, our analysis shows that FLAMe exhibits significantly lower bias compared to popular LLM-as-a-Judge models on the CoBBLEr autorater bias benchmark, while effectively identifying high-quality responses for code generation.
\section*{Limitations and Future work}
\label{sec:limitations}
Evaluating LLMs is challenging due to evolving evaluation standards and the need to assess new LLM capabilities. Expanding our data collection with open-source contributions could address this issue. Additionally, our models, trained primarily on English data with a context length of 2048 tokens, might not perform well on multilingual~\citep{freitag-etal-2021-experts} or long-context~\citep{kim-etal-2024-fables,karpinska-etal-2024-one} quality assessment tasks. In future releases, we plan to include training on more multilingual datasets with longer context lengths. Finally, in this work, we train our models in a supervised multitask fashion. Exploring alternative training approaches such as RLHF and DPO is a promising direction for future work.

\section*{Ethical Considerations and Risks}
All considerations and risks outlined by prior work for pretrained and instruction-tuned LLMs~\citep{chowdhery-etal-2022-scaling,anil-etal-2023-palm} apply to LLM autoraters. We recommend following standard practice for responsible development of these models~\citep{achiam-etal-2023-gpt,gemini-team-2023-gemini,reid-etal-2024-gemini}. Additionally, LLM autoraters raise new risks due to increased quality assessment capabilities. First, our models can inherit and amplify biases from human evaluations, leading to unfair or discriminatory outcomes. For instance, the model may replicate biases related to race, gender, or other sensitive attributes from the training data, potentially harming certain groups. Second, overreliance on LLM autoraters risks automating decisions that need human understanding and empathy. To mitigate these risks, transparency in model development and use, along with robust measures like bias audits, data anonymization, and incorporating diverse perspectives, is essential for promoting fairness, accountability, and trustworthiness.
\label{sec:ethics}
\section*{Acknowledgments}

We are grateful to Jie Ren, Denny Zhou, and Tania Bedrax-Weiss for their comments on this manuscript. We thank Mohit Iyyer, Daniel Cer, Elizabeth Clark, Jeremiah Liu, Balaji Lakshminarayanan, Clara Huiyi Hu, Aliaksei Severyn, Adam Sadovsky, Yonghui Wu, Quoc Le, Slav Petrov, Séb Arnold, Taylan Bilal, Noah Constant, Colin Raffel, Nan Hua, Marzena Karpinska, Yixiao Song, Tuhin Chakrabarty, the Gemini model quality team, the Descartes team at Google, and the UMass NLP group for useful discussions and valuable feedback at different stages of this project. We thank the authors of the datasets used in this work, especially Niklas Muennighoff, Hyungjoo Chae, Mounica Maddela, Tanya Goyal, and Yuanhao Wu, for their helpful suggestions and for answering our questions. Finally, we thank Grady Simon, Chung-Ching Chang, Sho Kannan, Gustavo Hernandez Abrego, and the T5X team for their assistance with the codebase, implementation, and computational resources.

\bibliographystyle{abbrvnat}
\nobibliography*
\bibliography{custom}

\newpage

\section*{Appendix}
\section{List of Training Datasets in FLAMe}
\label{appendix:training-data-list}
\tableref{tab:training-dataset-list} shows the list of datasets used in our study.

\section{Analyzing Length and Token Bias in RewardBench}
\label{appendix:rb-analysis}
In this section, we provide an analysis of length (\appendixref{appendix:length-bias-rewardbench}) and token (\appendixref{appendix:token-bias-rewardbench}) bias issues identified in the RewardBench benchmark. Given these issues, we encourage future work to evaluate LLM autoraters on a wide variety of benchmarks (such as our evaluation suite in \sectionref{sec:experiments}), rather than relying solely on RewardBench.

\subsection{Length Bias in RewardBench}
\label{appendix:length-bias-rewardbench}
\tableref{tab:rewardbench-bias} highlights length bias in RewardBench. Overall, RewardBench shows significant imbalance across categories regarding length: Chat Hard, Math, and Coding favor shorter outputs, while Chat leans towards longer outputs. An adversarial submission might strategically select longer or shorter outputs based on prompt categories to achieve higher scores, without necessarily reflecting a genuinely strong preference model.

\begin{table}[h]
\small
\centering
\begin{tabular}{lc }
\toprule
RewardBench Category & \% Preference for Longer Outputs \\
\midrule
Chat & 79.1\% \\
Chat Hard & 29.6\% \\
Math & 6.5\% \\
Coding & 35.7\% \\
Safety & 41.9\% \\
\bottomrule
\end{tabular}
\caption{A summary of length bias in RewardBench. Overall, we find that four out of five RewardBench categories show a strong preference towards either longer or shorter outputs.}
\label{tab:rewardbench-bias}

\end{table}

\subsection{Token Bias in RewardBench}
\label{appendix:token-bias-rewardbench}
Besides length bias, we identified token bias in the Math and Safety categories of RewardBench. In Safety, favored responses significantly leaned towards phrases like \emph{``I'm sorry''}, which suggest hedged responses. The word ``sorry'' appeared nearly 23\% more frequently in preferred responses compared to non-preferred ones. Similarly, the Math split exhibited token bias, where tokens such as ``i'', ``can'', ``need'', ``to'', ``find'' were predominantly found in rejected responses.

\begin{table*}[t!]
\scriptsize
\begin{center}
\begin{tabular}{ llll } 
 \toprule
Capability & Dataset & Source & Output Format \\
\midrule
General Response Quality 

& BeaverTails Helpfulness & \citet{ji-etal-2023-beavertails} & Pairwise \\
& HH RLHF Helpfulness & \citet{bai-etal-2022-training} & Pairwise \\
& Hurdles LFQA & \citet{krishna-etal-2021-hurdles} & Pairwise \\
& LMSYS Chatbot Arena conversations & \citet{zheng-etal-2023-judging} & Pairwise \\
& MAUVE & \citet{pillutla-etal-2021-mauve} & Pairwise \\
& News Summary Evaluation & \citet{goyal-etal-2022-news} & Pairwise \\
& PRD & \citet{li-etal-2024-prd} & Pairwise  \\
& SHP & \citet{ethayarajh-etal-2022-understanding} & Pairwise \\
& HelpSteer & \citet{wang-etal-2023-helpsteer} & Pairwise, Pointwise \\
& Summary Comparisons & \citet{stiennon-etal-2020-learning} & Pairwise, Pointwise \\
& GENIE & \citet{khashabi-etal-2022-genie} & Pairwise, Pointwise, Generative \\
& Fine-grained RLHF & \citet{wu-etal-2023-fine} & Pairwise, Classification \\
& InstruSum & \citet{liu-etal-2024-benchmarking} & Pairwise, Classification \\
& WebGPT & \citet{nakano-etal-2021-webgpt} & Pairwise, Generative \\
& LENS & \citet{maddela-etal-2023-lens} & Pointwise \\
& SummEval & \citet{fabbri-etal-2021-summeval} & Pointwise\\
& riSum & \citet{skopek-etal-2023-towards} & Pointwise, Classification \\
& FeedbackQA & \citet{li-etal-2022-using} & Pointwise, Generative \\
& CoLA & \citet{warstadt-etal-2019-neural} & Classification \\
& SEAHORSE & \citet{clark-etal-2023-seahorse} & Classification \\
& CREPE & \citet{yu-etal-2023-crepe} & Classification, Generative \\
& Scarecrow & \citet{dou-etal-2022-gpt} & Classification, Generative  \\
& Validity LFQA & \citet{xu-etal-2022-answer} & Classification, Generative \\
 \midrule
Factuality/Attribution 
& MOCHA & \citet{chen-etal-2020-mocha} & Pointwise \\
& Sentence Similarity - C$\times$C & \citet{parekh-etal-2021-crisscrossed} & Pointwise  \\
& Sentence Similarity - STS-B & \citet{cer-etal-2017-semeval} & Pointwise \\
& WikiBio Hallucination & \citet{manakul-etal-2023-selfcheckgpt} & Pointwise \\
& BEGIN & \citet{dziri-etal-2022-evaluating} & Classification \\
& DialFact & \citet{gupta-etal-2022-dialfact} & Classification \\
& FActScore & \citet{min-etal-2023-factscore} & Classification \\
& FRANK & \citet{pagnoni-etal-2021-understanding} & Classification \\
& FaithDial & \citet{dziri-etal-2022-faithdial} & Classification \\
& HaluEval & \citet{li-etal-2023-halueval} & Classification \\
& MNLI & \citet{williams-etal-2018-broad} & Classification \\
& MultiPIT & \citet{dou-etal-2022-improving} & Classification \\
& PAWS & \citet{zhang-etal-2019-paws} & Classification \\
& Q$^2$ & \citet{honovich-etal-2021-q2} & Classification \\
& QAGS & \citet{wang-etal-2020-asking} & Classification \\
& QQP & \citet{iyer-etal-2017-first} & Classification \\
& VitaminC & \citet{schuster-etal-2021-get} & Classification \\
& RAGTruth & \citet{wu-etal-2023-ragtruth} & Classification \\
& ESNLI & \citet{camburu-etal-2018-snli} & Classification, Generative \\
& XSum Hallucination & \citet{maynez-etal-2020-faithfulness} & Generative \\
\midrule
Mathematical Reasoning & PRM800K & \citet{lightman-etal-2024-lets} & Pairwise \\
\midrule
Coding & Code Contests & \citet{li-etal-2022-competition} & Pairwise \\
& COFFEE & \citet{moon-etal-2023-coffee} & Pairwise \\
& CommitPack & \citet{muennighoff-etal-2023-octopack} & Pairwise \\
& CommitPack - Bugs & \citet{muennighoff-etal-2023-octopack} & Pairwise \\
\midrule
Safety & BeaverTails Harmlessness & \citet{ji-etal-2023-beavertails} & Pairwise \\
& HH RLHF Harmlessness & \citet{bai-etal-2022-training} & Pairwise  \\
& HH RLHF Red Teaming & \citet{bai-etal-2022-training} & Pointwise \\
& BeaverTails QA-Classification & \citet{ji-etal-2023-beavertails} & Classification \\
\midrule
Instruction Tuning & LIMA & \citet{zhou-etal-2023-lima} & Generative \\
& PRM800K IF & \citet{lightman-etal-2024-lets} & Generative \\
& TULU-2 & \citet{ivison-etal-2023-camels} & Generative\\

\bottomrule
\end{tabular}
\end{center}
\caption{A complete list of training datasets in our FLAMe collection, including their output formats and categorized capabilities. We derive multiple tasks from certain datasets. For example, HelpSteer~\citep{wang-etal-2023-helpsteer} includes human annotations for different attributes of model responses such as Helpfulness, Correctness, Coherence, Complexity, and Verbosity, allowing us to create distinct tasks, each focused on a specific attribute.}
\label{tab:training-dataset-list}
\end{table*}

\end{document}